\documentclass{article}
\usepackage[linesnumbered,ruled,vlined]{algorithm2e}

\usepackage{arxiv}

\usepackage[utf8]{inputenc} 
\usepackage[T1]{fontenc}    
\usepackage{hyperref}       
\usepackage{url}            
\usepackage{booktabs}       
\usepackage{amsfonts}       
\usepackage{nicefrac}       
\usepackage{microtype}      
\usepackage{lipsum}		
\usepackage{graphicx}
\usepackage{natbib}
\usepackage{doi}
\usepackage{xcolor}
\usepackage{amsmath}
\usepackage{xcolor}
\usepackage[normalem]{ulem}

\title{Unsupervised Atomic Data Mining via Multi-Kernel Graph Autoencoders for Machine Learning Force Fields}

\author{
  Hong Sun\thanks{These authors contributed equally to this work}\\
  \href{https://orcid.org/0000-0000-0000-0000}{\includegraphics[scale=0.06]{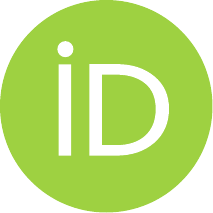}}\hspace{1mm}Physics Division\\
  Lawrence Livermore National Laboratory\\
  Livermore, CA 94551\\
  \texttt{sun36@llnl.gov}
\And
  Joshua A. Vita\footnotemark[1]\\
  Materials Science Division\\
  Lawrence Livermore National Laboratory\\
  Livermore, CA 94551\\
  \texttt{vita1@llnl.gov}
\And
  Amit Samanta\\
  Physics Division\\
  Lawrence Livermore National Laboratory\\
  Livermore, CA 94551\\
  \texttt{samanta1@llnl.gov}
  \And
  Vincenzo Lordi\thanks{Corresponding author}\\
  Materials Science Division\\
  Lawrence Livermore National Laboratory\\
  Livermore, CA 94551\\
  \texttt{lordi2@llnl.gov}
}

\renewcommand{\shorttitle}{}

\hypersetup{
pdftitle={A template for the arxiv style},
pdfsubject={q-bio.NC, q-bio.QM},
pdfauthor={David S.~Hippocampus, Elias D.~Striatum},
pdfkeywords={First keyword, Second keyword, More},
}

\begin{document}
\maketitle

\newcommand{\comment}[1]{\begingroup\color{red}#1\endgroup}

\begin{abstract}

Constructing a chemically diverse dataset while avoiding sampling bias is critical to training efficient and generalizable force fields.
However, in computational chemistry and materials science, many common dataset generation techniques 
are prone to oversampling regions of the potential energy surface.
Furthermore, these regions can be difficult to identify and isolate from each other or may not align well with human intuition, making it challenging to systematically remove bias in the dataset.
While traditional clustering and pruning (down-sampling) approaches can be useful for this, they can often lead to information loss or a failure to properly identify distinct regions of the potential energy surface due to difficulties associated with the high dimensionality of atomic descriptors.
In this work, we introduce the Multi-kernel Edge Attention-based Graph Autoencoder (MEAGraph) model, an unsupervised approach for analyzing atomic datasets.
MEAGraph combines multiple linear kernel transformations with attention-based message passing to capture geometric sensitivity and enable effective dataset pruning without relying on labels or extensive training.
Demonstrated applications on niobium, tantalum, and iron datasets show that MEAGraph efficiently groups similar atomic environments, allowing for the use of basic pruning techniques for removing sampling bias.
This approach provides an effective method for representation learning and clustering that can be used for data analysis, outlier detection, and dataset optimization.

\end{abstract}


\section{Introduction}

In the domain of atomistic simulations, machine learning force fields (MLFFs) are becoming vital tools for large-scale simulations of chemical reactions and material properties, often achieving accuracies comparable to much more expensive quantum mechanical calculations upon which they are trained \cite{behler2007generalized, schutt2017quantum, bartok2010gaussian,bartok2017machine}. The development of an MLFF can be framed as a regression problem, where the goal is to reproduce the complex non-convex potential energy surface (PES) by training on atomic datasets sampled from a wide range of atomic environments \cite{deringer2021gaussian, rupp2012fast,behler2014representing}. 
Constructing an optimal training dataset for an MLFF  is a challenging process, often involving human intuition and intervention, as well as the costly task of assigning corresponding labels (e.g., energies and forces) to the data using electronic structure calculations such as density functional theory (DFT).
In addition, the ``expert intuition'' guided approach of selecting atomic configurations leads to uncontrolled systematic errors and bias, as well as imprecise control over the number of configurations, atoms per structure, and the associated representative sampling of atomic environments.
Other popular approaches, such as use of molecular dynamics simulations or perturbed crystal structures to seed datasets, often lead to highly-correlated data samples and systematic redundancy.
As a result, these datasets often suffer from data imbalances including an over-representation of low-energy equilibrium data due to the influence of Boltzmann statistics in molecular dynamics (MD) simulations and an under-sampling of complex atomic environments, such as those involving non-equilibrium dynamics or defects.
Therefore, to improve the performance of MLFFs and decrease the cost of DFT-based data preparation, it is crucial to develop effective data mining techniques to build high-quality atomic datasets.

Recent studies on data pruning techniques \cite{scalinglaw, nguyen2022quality, rosenfeld2021predictability} challenge the notion that larger models and datasets are always better, suggesting instead that datasets can be carefully sub-sampled to yield smaller training sets that still yield excellent model performance.
The key insight is that training datasets typically contain redundant data points \cite{cherti2023reproducible, tirumala2023d4} that can be safely removed without losing any critical information.
By applying high-quality data pruning methods to remove these redundancies, models can be trained on smaller optimized datasets without compromising on diversity or performance, leading to more efficient training of deep learning models.
Indeed, the use of such smaller, optimized datasets can significantly improve training by avoiding parameter optimization into local minima.
The importance of effective data pruning and appropriate data sampling is increasingly recognized in practical applications in diverse fields, including computer vision, natural language processing, speech recognition, and scientific domains such as computational chemistry and materials science \cite{scalinglaw, azeemi2023data,metadynamics,toneva2019empirical, paul2022deep}.

The success of pruning and data selection techniques relies on the underlying assumption that the initial dataset is sufficiently diverse and the data selection techniques can help in decreasing existing redundancies.
To generate a diverse dataset, researchers have explored many data sampling techniques including entropy optimization \cite{entropyoptim,quests}, metadynamics sampling \cite{metadynamics}, and active learning \cite{kulichenko2023uncertainty,schwalbe2021differentiable,zaverkin2022exploring}.
Although these strategies are promising, they also have limitations. Constructing a robust training set through entropy optimization necessitates striking a balance between entropy-maximized and energy-minimized configurations. Active learning relies on context-dependent hyperparameters, and free energy sampling methods like metadynamics require manual selection of collective variables. 
Using a combination of techniques can be advantageous as well.
A secondary problem is that MLFFs are typically trained on per-atom quantities (such as forces) and total energies, but most data sampling and pruning strategies focus on selecting or removing entire atomic structures, each containing tens or hundreds of atoms{\cite{entropyoptim,metadynamics,kulichenko2023uncertainty,schwalbe2021differentiable,zaverkin2022exploring}}; 
this approach can create challenges in maintaining a balanced representation of a training set in the atomic feature space, since many similar atomic environments may be present in a structure used to represent only a small number of unique ones.

In this study, we introduce a novel approach for unsupervised data mining of atomic datasets based on a multi-kernel edge attention graph autoencoder, which we call MEAGraph (Multi-kernel Edge Attention-based Graph Autoencoder).
By using a message-passing framework and attention mechanism, our model learns robust representations of atomic environments with minimal hyperparameters, allowing it to provide a comprehensive assessment of atomic environment similarity that leverages long-range information and is more dynamic than simple Euclidean distance.
We test this MEAGraph model to analyze three atomic datasets---niobium (Nb), tantalum (Ta), and iron (Fe)---with the following objectives: (1) to identify atomic environments in complex configurations by clustering atoms and identifying anomalies/uniqueness, (2) to prune atomic data by removing  redundant environments 
from within each cluster and evaluating the performance of a force field trained on the pruned datasets, and (3) to investigate how individual clusters impacts the performance of MLFFs.

The main contributions of this work include:

\begin{itemize}
    \item The development of MEAGraph, a novel multi-kernel edge attention graph autoencoder capable of learning robust representations of atomic environments for use with similarity analysis and clustering.
    \item Qualitative analysis showing good agreement between clusters identified using MEAGraph and expectations based on known physics and ground-truth force distributions, outperforming existing methods.
    \item Demonstrations applying MEAGraph for detecting and removing redundancy in an unsupervised manner on multiple example datasets.
    \item Discussion of the value of cluster-based analysis of datasets and highlighting the limitations of our technique.
\end{itemize}

MEAGraph accurately identifies both similar and structurally different atomic environments, enabling the construction of optimized, more compact datasets for improved training efficiency and less biased models.
The edge attention graph reduction approach allows MEAGraph to learn compact representations of atomic environments irrespective of the size of the feature space.
This flexibility makes the approach highly adaptable for applications such as clustering-based outlier detection and efficient data pruning in high-dimensional feature spaces. 

\section{Method}
\subsection{MEAGraph model}

\begin{figure}
    \begin{center}
        \includegraphics[width= 0.95\columnwidth]{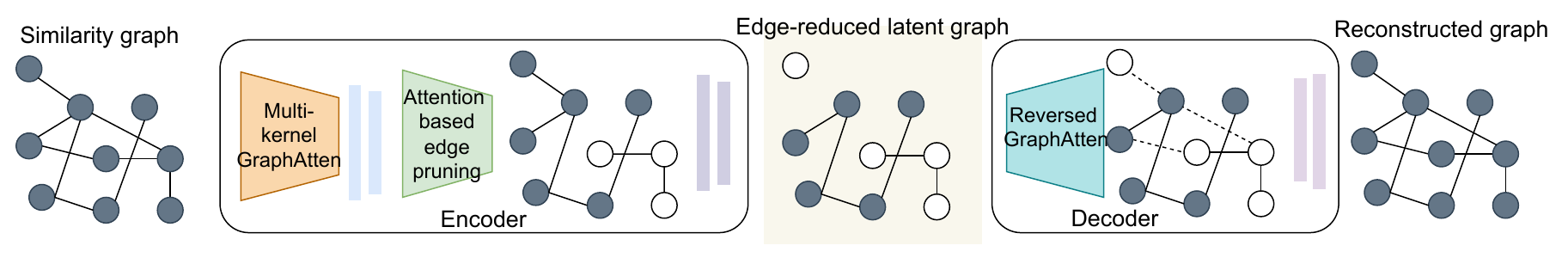}
    \end{center}
    \caption{Schematic illustration of the MEAGraph model. First, a similarity graph is built with nodes representing atoms and edges defined by feature similarity. Next, this similarity graph is used by the encoder to generate a compact edge-reduced graph. The encoder consists of multiple graph attention layers that update node features using linear kernel transformations and aggregate messages through edge attention. During training, edges with low similarity scores are pruned at each layer and later restored using reversed graph attention layers so as to minimize the reconstruction loss between the original and reconstructed feature matrices.}
    \label{fig:schematics}
\end{figure}

The general architecture of MEAGraph is described in Figure ~\ref{fig:schematics}, where graph attention layers are trained to preserve only the edges which are most valuable for reconstructing the original feature matrix. Depending upon the pooling rate used to set the degree of edge reduction, the latent graph generated by the model may be composed of multiple disjoint sets of points. The formation of these disjoint sub-graphs is critical to the applications of MEAGraph described in the remainder of this paper, where the assumption is made that points in the same sub-graph will have high similarity and can thus be treated as equivalent to each other for the purposed of environment identification and pruning.

Algorithm \ref{alg:build_graph} describes the construction of the similarity graph from input data composed of atomic feature vectors, such as Smooth Overlap of Atomic Positions (SOAP) \cite{soap}, Atom-Centered Symmetry Functions (ACSF) \cite{acsf}, bispectrum descriptors \cite{fitsnap}, or latent features learned on-the-fly in graph neural network (GNN) models such as NequIP \cite{nequip} and MACE \cite{mace}. In this graph, each node corresponds to an atom, and edges are connected between atoms based on the similarity of their feature vectors. The similarity score between two atoms, with feature vectors $x_i$ and $x_j$, is computed as the exponential of the negative Euclidean distance between them. To retain only the most relevant connections and reduce the size of the input graph, edges are preserved if their similarity score exceeds a predefined threshold $r_l$, thereby ensuring that only highly similar atomic environments are connected.

\begin{algorithm}
\caption{Build Graph}
\label{alg:build_graph}
\SetAlgoNlRelativeSize{-2}
\SetKwInput{KwInput}{Input}
\SetKwInput{KwOutput}{Output}
\KwInput{Atomic feature vectors  $X_0$, threshold $r_l$ for the normalized similarity score $\hat{s}_{ij}$ to determine edge formation}
\KwOutput{Output features $X$, edge index $E$}
$s_{ij} \gets \exp(-\| x_i - x_j \|)$ for $x_i, x_j \in X_0$, where $i \neq j$\;
$\hat{s} \gets \text{MinMaxNorm}(s)$\; 
$E \gets \{ (i, j) \mid \hat{s}_{ij} > r_l \}$\;
$X \gets \{ x_i \mid \exists j, (j, i) \in E \}$\;
\end{algorithm}

The training process of the MEAGraph model is detailed in Algorithm \ref{alg:MEAGraph} and involves iterative learning by using batches of input feature matrices $X$. In each iteration, the algorithm begins by constructing a similarity graph from the input features using the Build\_Graph function (Algorithm \ref{alg:build_graph}), then using an encoder to prune and pool edges. The encoder component consists of multiple GNN layers. Within each GNN layer, node (atom) features are updated through a message-passing process, which is followed by aggregation of messages from neighboring nodes. This aggregation is weighted by attention scores, which are calculated based on the Euclidean distances between the feature vectors of the nodes. These attention scores determine the relevance of neighboring nodes during feature updates, with higher scores indicating more relevant or similar atoms.
After each GNN layer, edge pooling is applied to simplify the graph structure. This involves removing edges with attention scores below a specified threshold $r$, which reduces the number of connections in the graph while retaining the most relevant relationships between nodes. The resulting pooled graph becomes more compact as it (iteratively) focuses only on essential interactions (similarities) between atomic environments in the feature space. 

\begin{algorithm}
\caption{MEAGraph training}
\label{alg:MEAGraph}
\SetKwInput{KwInput}{Input}
\KwInput{ Feature matrix $X$, Edge index $E$, Number of training iterations $T$, Number of batches $B$, Number of layers $L$, Number of kernels $K$, and edge pooling rate $r$.}

\For{$t \leftarrow 1$ \KwTo $T$}{
    \For{$b \leftarrow 1$ \KwTo $B$}{
        \text{edge\_index } $E_{b} \gets \text{Build\_Graph}(X_{b})$\;
        \textbf{Encoder:}\\
        \For{$l \leftarrow 0$ \KwTo $L-1$}{
            \For{$k \leftarrow 1$ \KwTo $K$}{
                \textbf{GNN Layer:}\\
                $\hat{h}_{l}^{k} = \text{BatchNorm}( W_{1}^{k} h_{l}^{k})$\;
                $\alpha_{ij,l}^{k} = \frac{\exp\left(-\beta^{k} \cdot \| \hat{h}_{i,l}^{k} - \hat{h}_{j,l}^{k}\| \right)}{\sum_{j=1}^{n} \exp\left(-\beta^{k} \cdot \| \hat{h}_{i,l}^{k} - \hat{h}_{j,l}^{k} \|\right)}$\;
                $h_{i,l+1}^{k} = \sigma \left( W_{1}^{k} h_{i,l}^{k} + W_{2}^{k} \cdot \text{Agg}_{j \in N(i)} \left( \alpha_{ij,l}^{k} \cdot W_{1}^{k} h_{j,l}^{k} \right) \right)$\;
            }
            $h_{l+1} = \text{mean}_{k \in 1...K} h_{l+1}^{k}$\;
            $\alpha_{l+1} = \text{mean}_{k \in 1...K} \alpha_{l+1}^{k}$\;
            $\hat{\alpha}_{l+1} = \text{MinMaxNorm}(\alpha_{l+1})$\;
            $\tilde{E}_{b,l+1} \gets \left\{ e \mid e \in E_{b,l+1}, \hat{\alpha}_{l+1}(e) > r \right\}$\;
        }
        
        \textbf{Decoder:}\\
        \For{$l \leftarrow L-1 $ \KwTo $0$}{
            $\tilde{h}_{l} \gets GNN(\tilde{h}_{l+1}, \tilde{E}_{b,l+1})$\;
        }
        $\tilde{X}_{b} \gets \mathbb{W} \cdot \tilde{h}_{0} + B$\;  
        
        $L \gets \mathbb{E} \left[ \|\tilde{X}_{b} - X_{b}\|^2 \right]$\;
        ${\theta} \gets \text{Optim}(L, \theta)$\;
    }
}
\end{algorithm}

The decoder part of the model reconstructs the node features from the pruned graph by applying an inverse of the GNN layers used in the encoder, restoring the original feature matrix as closely as possible. The model is trained by minimizing the reconstruction loss of the decoder, which is the difference between the original input features and the reconstructed features. Through this process, the model learns to capture the most relevant information in the atomic dataset while discarding unnecessary details. The MEAGraph model's edge-aware attention mechanism and edge pooling technique effectively groups atoms with similar atomic environments, allowing the removal of redundant information within these groups and resulting in a more compact, informative representation of the input data.
For the results presented in this manuscript, the hyperparameters of the MEAGraph model are described in Table ~\ref{tab:hyperparams}.


\begin{table}
    \centering
    \label{tab:hyperparams}
    \caption{Critical hyperparameters of the MEAGraph models used in this work. See Algorithms ~\ref{alg:build_graph} and ~\ref{alg:MEAGraph} for descriptions of symbols. Note that in all cases an adaptive pooling rate was used during training, where values are randomly sampled in the range [0,1] for each batch so that the model could be safely used at any pooling level during inference. The $r$ values reported in this table are the values used during inference for each dataset.}
    \begin{tabular}{cccccccc}
        \hline \hline
        Dataset & Descriptor & $r_l$ & $r$ & $L$ & $K$ & $E$ & $B$ \\
        Nb & SOAP & 0.8 & 0.3, & 2 & 6 & 50 & 1 \\
        Ta & Bispectrum & 0.9 & 0.9 & 2 & 6 & 600 & 4 \\
        Fe & Bispectrum & 0.9 & 0.7 & 2 & 6 & 20 & 8 \\
        \hline

        
        \hline \hline
    \end{tabular}
\end{table}

\subsection{Force field training}

We demonstrate the effectiveness of our data mining approach discussed above
by training a MEAGraph model to the full training set, classifying similar atomic environments into groups, pruning redundant atoms, then training a force field to the remaining atoms. 
We apply this technique to three different published elemental datasets, Ta \cite{Ta_dataset} and Fe \cite{Fe_dataset}, for the purpose of dataset pruning.
%
%
In addition, we studied the clustering performance of MEAGraph for the analysis and identification of atomic environments for an additional elemental dataset for Nb \cite{sun2023Nb_GB}.
We trained solely on forces, and produced force fields that predict forces only, using
per-atom bispectrum descriptors generated using the FitSNAP program \cite{fitsnap} as input. These bispectrum descriptors, along with the corresponding force labels for each atom in all three directions, are combined to form the feature matrix and label matrix, which are used to train a ridge regression based force field model. The dataset is randomly split into training and test sets with a ratio of 80\% for training and 20\% for testing. To ensure a fair evaluation, the test set is balanced by equalizing the number of atoms across each configuration type for the labeled Ta and Fe datasets, allowing for an unbiased assessment of model performance.

\section{Results and Discussion}
\label{sec:others}

\subsection{Atomic environment identification:  Nb dataset}
\label{sec:Nb}

\begin{figure}
    \begin{center}
        \includegraphics[width= 0.9\columnwidth]{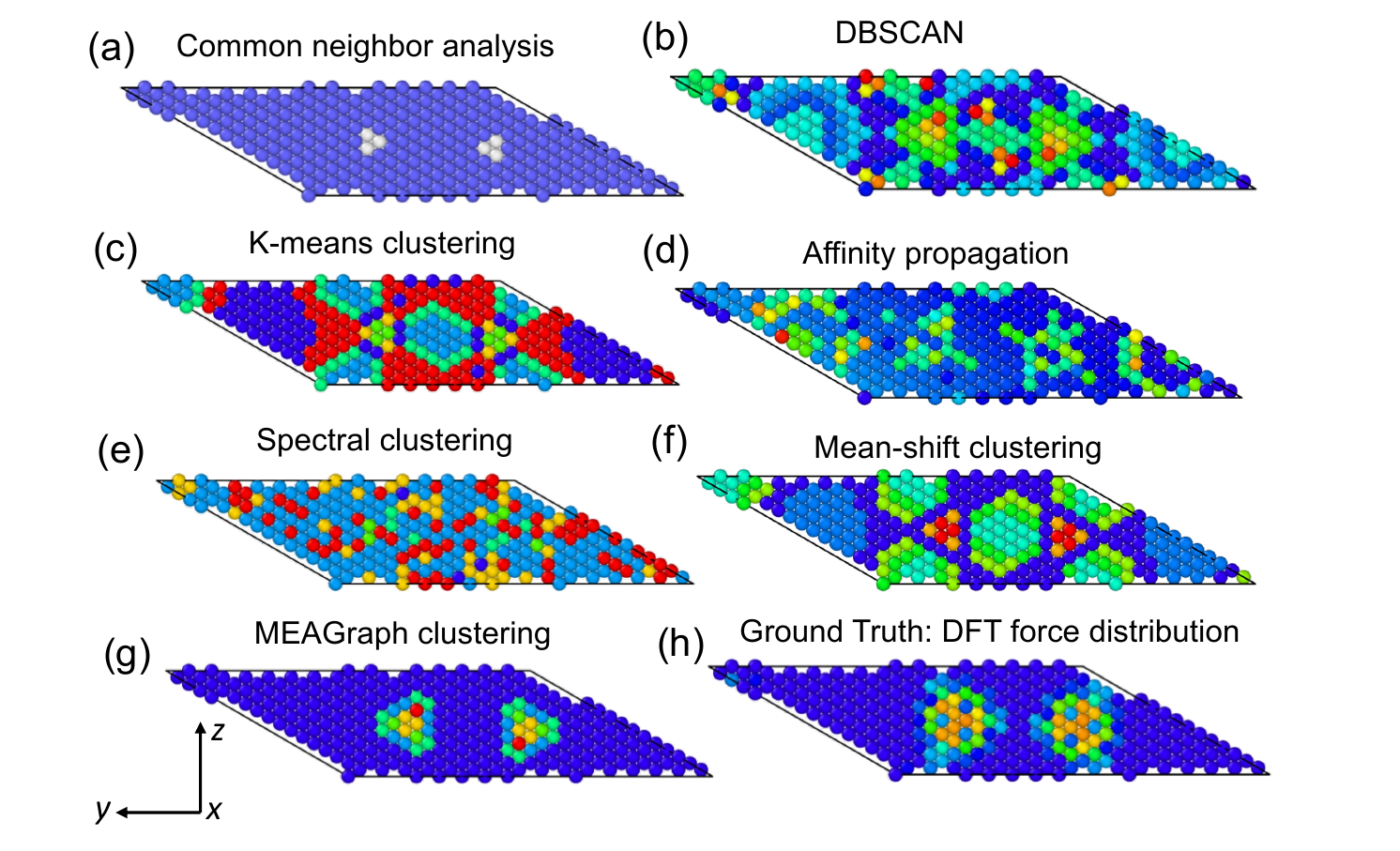}
    \end{center}
    \caption{Identification of atomic environments in the easy-hard split dislocation core configuration of the Nb system using common neighbor analysis and clustering methods based on per-atom SOAP descriptors. Among the clustering methods, only MEAGraph successfully identifies atomic environment groups that align with those indicated by the ground truth DFT force distribution. In panels (a-g), the colors represent the atomic environment groups identified by each clustering method, while in panel (h), the colors represent the magnitudes of the DFT forces.}
    \label{fig:clustering}
\end{figure}

Accurately identifying atomic phases or environments is crucial for understanding the structure-property evolution of materials during MD simulations. In simple systems, such as face-centered cubic (FCC) or body-centered cubic (BCC) metals, traditional structural analysis methods like common neighbor analysis (CNA) or bond-orientational order parameters \cite{Steinhardt1983} are often used to study phase evolution. However, identification of more complex atomic phases or environments typically require applying clustering techniques to atomic descriptors, due to classification uncertainties or emergent structures \cite{clustering_gold,clustering_phosphorus,clustering_amorphous}. Selecting an appropriate clustering method is therefore essential for the accurate classification of atomic environments, particularly for automated analysis suitable for on-the-fly analysis of large MD trajectories. Figure~\ref{fig:clustering}(a-g) compares the results of 7 different clustering methods, including MEAGraph, for identifying and tagging different atomic environments  in a test simulation cell containing split screw dislocation cores in a Nb crystal with a BCC structure. The particular dislocation structure is a split easy- and hard-core configuration with Burgers vector and dislocation line direction parallel to the $[111]$ direction and was obtained from ref.\cite{sun2023Nb_GB}. For this analysis, SOAP descriptors were used to encode the atomic environments in the structure, with the descriptor hyperparameters set to $r_{cut}=5.0$, $n_{max}=3$, and $l_{max}=3$.\cite{soap} 

The clustering results are compared to the force magnitude distribution obtained from DFT calculations, shown in Fig.~\ref{fig:clustering}(h), which clearly shows a distinction between a bulk region mostly showing near-zero forces (depicted in blue) and the two dislocation core regions with similar color patterns that show the three-fold rotational symmetry about the dislocation axis. Standard CNA identifies only two environment types: dislocation cores (white) and bulk atoms (blue). In contrast, popular clustering methods such as DBSCAN \cite{dbscan}, {\it k}-means \cite{kmeans}, spectral clustering \cite{spectral}, and affinity propagation \cite{affinity_prop} reveal more detailed and disparate atomic environments (the hyperparameter settings for each clustering method can be found in the code repository associated with this work). However, the patterns identified by these methods deviate from the patterns evident from the distribution of DFT forces on the atoms. On the other hand, the coloring of atoms based on clusters identified by the latent graph of MEAGraph around the dislocations closely resembles the pattern evident from the distribution of DFT forces on atoms, despite only structural information (no forces) being used by MEAGraph. The success of MEAGraph in accurately identifying the atomic environments without introducing excessive noise in the categorization suggests great utility for a range of materials science applications, including phase identification and defect analysis.

\subsection{Data pruning of Ta dataset}
\label{sec:Ta}

\begin{figure}
    \begin{center}
        \includegraphics[width=1\columnwidth]{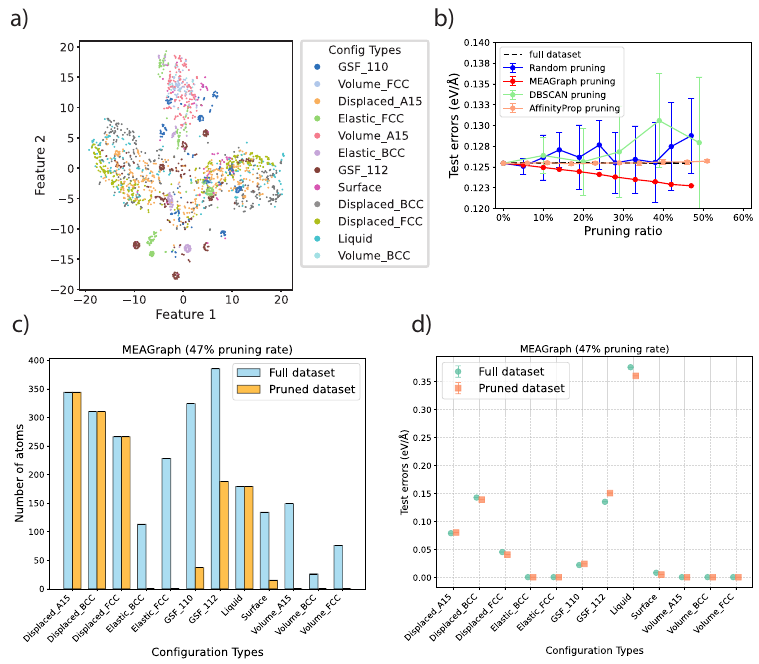}
    \end{center}
    \caption{(a) 2D t-SNE plot of the per-atom bispectrum descriptors of forces in the Ta dataset, with different configuration types distinguished by color. (b) Force prediction performance comparison on balanced test datasets, evaluated using four pruning methods: MEAGraph, DBSCAN, affinity propagation, and random pruning. Error bars indicate variations across 20 random iterations of the pruning procedures. (c) Distribution of atom counts in the original full dataset compared to the pruned dataset after applying MEAGraph pruning at a 47\% pruning ratio. (d) Test error comparison per configuration type between the full dataset and the pruned dataset at 47\% pruning ratio with MEAGraph method.}
    \label{fig:pruning_Ta}
\end{figure}

Sampling a diverse set of atomic environments is important to accurately train a MLFF for simulating complex material processes or explore broad regions of the non-convex potential energy landscape. Datasets used to train MLFFs are often generated by sampling atomic structures in near-equilibrium ensembles at various conditions (e.g., temperatures and pressures), often generated via molecular dynamics simulations, Monte Carlo simulations, or a combination of both), followed by resource intensive DFT calculations to obtain the training labels (i.e., energies and forces). Due to the high computational cost associated with dense sampling of the high-dimensional feature space, most datasets are constructed using expert intuition, which often results in non-uniform sampling of the feature space, while reliance on techniques such as MD or MC results in high redundancy and oversampling of near-equilibrium structures (e.g., slightly perturbed crystal environments). 
Correspondingly, typical hand-crafted training datasets often include many redundant data points which can introduce bias into the training process, leading to under/over-fitting of certain regions of the energy landscape over others. However, redundancy and implicit bias in a dataset are often difficult to detect from the feature patterns of atomic environments within the dataset. For example, Fig.~\ref{fig:pruning_Ta}(a) depicts the two-dimensional t-SNE feature map of bispectrum descriptors for per-atom forces in the Ta dataset from Ref.~\cite{Ta_dataset}, which includes 12 different types of structures, including bulk FCC and BCC structures with hydrostatic strains ($<$ 3\%), bulk structures with atoms perturbed randomly from their ideal sites by up to 3\% of the equilibrium lattice parameter, structures containing low-index surfaces, and bulk liquid structures obtained from MD simulations performed at temperatures above the melting point. The feature map in Fig.~\ref{fig:pruning_Ta}(a) does not reveal a distinct distribution pattern based on similarity of atomic features, with atoms from various configurations exhibit overlapping feature ranges, and thus is not suitable for  pruning of redundant data. Our conjecture is that a practical strategy to perform unsupervised data pruning without using DFT labels or a trained model is to cluster the dataset to groups, each consisting of atoms with similar features, and then prune the data within each cluster. To this end, it is important to use a method such as MEAGraph that can learn the geometry of the data in a most informative manner.

Figure~\ref{fig:pruning_Ta}(b) presents the root mean square errors (RMSE) between the predicted and true force labels in a test dataset for  MLFFs trained using different pruning strategies and comparing to MEAGraph. The MLFF models are trained with the FitSNAP package \cite{fitsnap} by using ridge regression with per-atom bispectrum descriptors for the forces across datasets pruned at different ratios (i.e., the fraction of atoms removed relative to the total number of atoms). In this section, the bispectrum descriptors were computed with hyperparameters $j_\text{max}=3$ and $r_\text{cutfac}=4.67$~\AA. To understand the importance of learning the intrinsic geometry of the data and to demonstrate the ability of MEAGraph to accurately identify similar and different environments, we have compared four pruning methods: MEAGraph, DBSCAN \cite{dbscan}, affinity propagation pruning \cite{affinity_prop}, and random pruning. For the three clustering-based pruning methods (i.e., MEAGraph, DBSCAN and affinity propagation), we first apply the respective clustering technique on the initial dataset, then prune the training dataset by randomly removing a certain fraction (ranging from 0\% to 100\%) of atoms from each cluster containing more than a predefined minimum threshold of atoms (set to 20, to avoid removing outlier points). For the case of random pruning, a total number of atoms are randomly removed from the entire training dataset, corresponding to the same number of atoms removed by MEAGraph for each pruning ratio. Since each clustering method creates a different cluster size distribution, the exact number of pruned atoms for a given pruning ratio is slightly different for MEAGraph, DBSCAN, and affinity propagation.

To ensure a fair comparison and eliminate any implicit bias in the original training dataset, we used a two-fold approach:  first, a balanced test dataset was obtained by equalizing the number of atoms across different configuration types, ensuring equal representation of each configuration type during model evaluation; second, we performed 20 random sampling iterations for each pruning ratio for all four methods. As shown in Fig.~\ref{fig:pruning_Ta}(b), the test errors for random pruning and DBSCAN tend to increase as the pruning ratio increases, while those for affinity propagation stay largely stagnant. Random pruning and pruning of clusters obtained from DBSCAN both also exhibit large variations (uncertainty) in results across the 20 iterations, whereas pruning of clusters obtained by affinity propagation results in lower mean test errors with minimal variations. In contrast, pruning of clusters obtained from MEAGraph leads to a monotonic decrease in test errors as the pruning ratio increases, with negligible uncertainties (error bars in Fig.~\ref{fig:pruning_Ta}(b) are smaller than the symbols for MEAGraph). The mean RMSE values for pruning using MEAGraph are close to the lower bounds obtained from random pruning and are achieved with only 1 iteration. (Since very little run-to-run variation is observed with MEAGraph, it is more efficient than the other methods in not requiring multiple random instantiations to find the minimum-error converged result.) These results demonstrate that the MLFF performs better on a balanced test dataset when MEAGraph pruning is applied, as it effectively and reliably removes redundant atoms from large clusters with similar atomic environments. Notably, for this test case, up to 47\% of the atoms can be pruned with MEAGraph while achieving even lower RMSE values compared than the original, unpruned dataset.

\newcommand{\grouplabel}[1]{\texttt{#1}}

Figure~\ref{fig:pruning_Ta}(c) shows the distribution of atoms across different human-labeled groups in the full dataset, along with the distribution after 47\% pruning using MEAGraph clustering. The differences of the two distributions indicates where MEAGraph detected meaningful redundancies and also further illustrates the challenges of identifying redundacy using expert intuition, since the largest discrepancies (e.g., \grouplabel{Displaced\_A15} vs \grouplabel{Volume\_A15}) may be counterintuitive based on the group labels. The clustering process identified the most redundancy in structures generated from elastically (hydrostatic or shear) strained crystal structures, namely the groups labeled \grouplabel{Elastic\_BCC}, \grouplabel{Elastic\_FCC}, \grouplabel{Volume\_A15}, \grouplabel{Volume\_BCC}, and \grouplabel{Volume\_FCC}, where the entire clusters were removed during pruning. Additionally, nearly 90\% of atoms in the \grouplabel{GSF\_110} group (i.e., structures containing stacking faults on the $[111]$ plane in the $(111)$ direction), $\sim$35\% in the \grouplabel{GSF\_112} group   (i.e., structures containing stacking faults on the $[111]$ plane in the $(112)$ direction), and $\sim$50\% of atoms in the \grouplabel{Surface} (i.e., structures containing low-index surfaces) were pruned as redundant. In contrast, other groups, such as \grouplabel{Displaced\_A15}, \grouplabel{Displaced\_BCC}, and \grouplabel{Displaced\_FCC} (i.e., structures in which atoms were randomly perturbed off-lattice for different crystal structures), as well as \grouplabel{Liquid},  were found to be contain mostly unique environments and were left largely intact by the pruning process. These results suggest that the atomic environments in these latter groups are unique and may be essential for training a force field model with well-sampled, independent atomic configurations, at least for regions of the energy landscape close to the respective structures. (We note as an aside, that in this case, the transferability of such a trained model is quite good, as the random configurations in the \grouplabel{Displaced} and \grouplabel{Liquid} groups, along with the few additional examples in the highly-redundant groups, span a good representation of the configuration space of Ta for many simulation conditions.) 

The efficacy and performance of the retrained MLFF using the highly pruned dataset from MEAGraph clustering is highlighted in 
Figure~\ref{fig:pruning_Ta}(d), which compares the test errors averaged over the 20 pruning runs for each labeled group for the force field trained using either the full dataset or the 47\% MEAGraph pruned one, showing similar or lower test errors after pruning, with only slightly increased test errors for the \grouplabel{GSF} groups. With test errors generally decreasing as the pruning ratio increases shown in Figure~\ref{fig:pruning_Ta}(b) and comparable or reduced per-group test errors compared to the full dataset seen in Figure~\ref{fig:pruning_Ta}(d), we can observe that MEAGraph pruning effectively removes redundancy and bias from the training dataset, providing improved (more uniform) dataset sampling, in a manner that maintains or improves force field predictions, while also significantly reducing the training complexity. Notably, the pruning process is based entirely on clustering analysis performed on the feature space of the original dataset, without requiring DFT labeling or pre-training.

\subsection{Cluster-based dataset analysis}
\label{sec:Fe}

In the previous section, we showed how randomly pruning a uniform percentage of atoms from every cluster obtained by MEAGraph (above a certain threshold size) was effective at removing redundancy from labeled groups in the dataset.
The assumption with this approach was that atoms which are in the same cluster are redundant with each other, meaning they can be safely pruned without negatively affecting the test error or even improving the test error by removing cluster redundancy (improving data sampling) during the training.
While this appears to be the case with the Ta dataset, it is possible for a cluster to lack a sufficient number of training points to adequately learn the forces of those atoms, thus leading to an increased test error if pruning is performed, particularly for small clusters and/or sparse clustering.
To probe this idea, we study the effects of pruning clusters individually, using the Fe dataset from \cite{Fe_dataset}, which is a more realistic example of a typical training set due to its larger size and higher structural diversity.

Following the same approach used for clustering the Ta dataset, we generated per-atom bispectrum descriptors of the force components for the Fe dataset using the FitSNAP package. For this dataset, the bispectrum descriptors were computed with hyperparameters $j_\text{max}=3$ and $r_\text{cutfac}=4.52$~\AA. These descriptors, along with the force labels from Ref.~\cite{Fe_dataset}, served as input to train a ridge regression model to predict atomic forces. As in the analysis for the Ta dataset, test errors were evaluated by using a balanced test set containing an equal number of atoms from each configuration type in the dataset.
MEAGraph clustering was then applied to the feature space of per-atom forces of the entire training dataset.
Finally, we computed pruning curves by randomly removing a fraction of points from one MEAGraph-determined cluster while leaving all other clusters unchanged and performed this procedure for each of the clusters individually. Clusters containing only one or two atoms were not pruned, since we observed that pruning these consistently resulted in higher test errors, demonstrating that these ``outlier'' clusters have no inherent redundancy with the others and were properly identified by MEAGraph as being dissimilar from the remaining pool of configurations, thus representing important independent information that should not be removed from training.

\begin{figure}
    \begin{center}
        \includegraphics[width=\columnwidth]{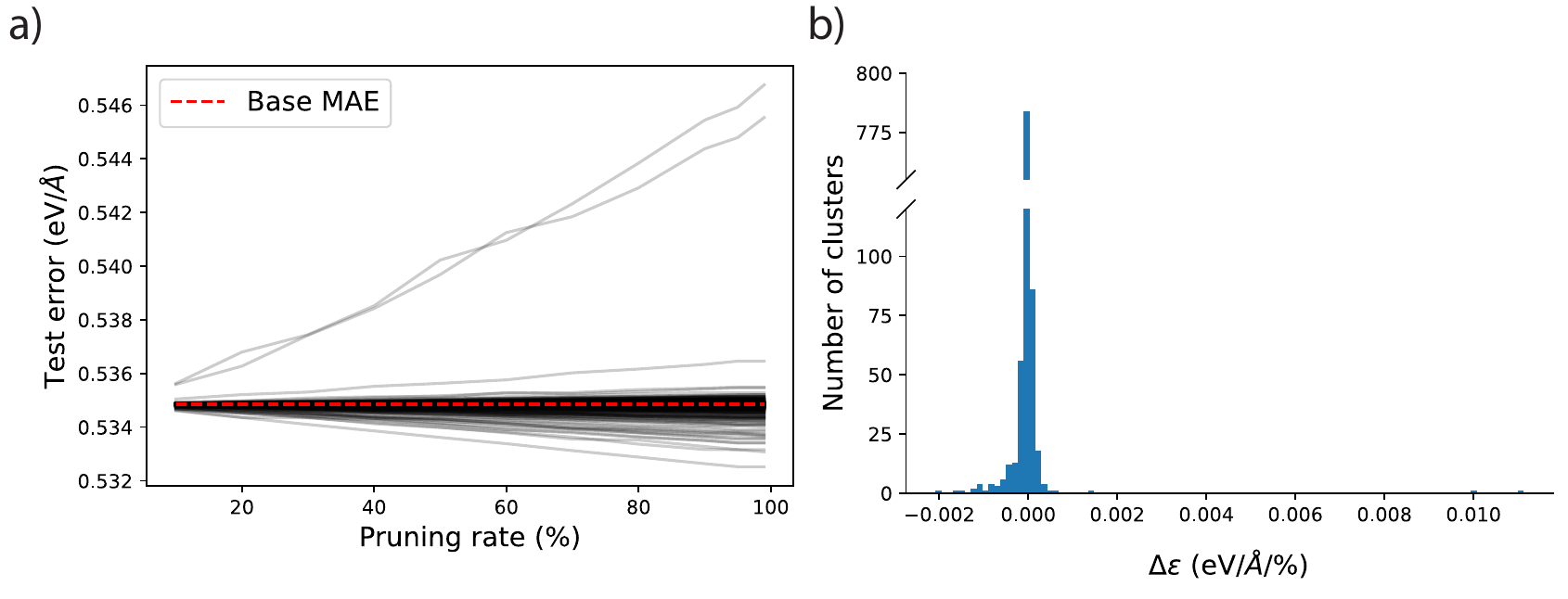}
    \end{center}
    \caption{(a) Pruning curves for the Fe dataset generated by individually pruning each cluster obtained by MEAGraph containing more than two atoms, where each line is the pruning curve for a separate cluster. We observe that pruning from the vast majority of clusters improves the test error, while a few clusters sharply increased test error upon pruning them. (b) The histogram of the average slopes of these curves shows that, although $\sim 66\%$ of clusters have a slope that is less than or equal to 0, a small number of clusters lead to an increased test error when pruned (positive $\Delta$ test error).
    }
    \label{fig:pruning_Fe}
\end{figure}

\begin{figure}
    \begin{center}
        \includegraphics[width=\columnwidth]{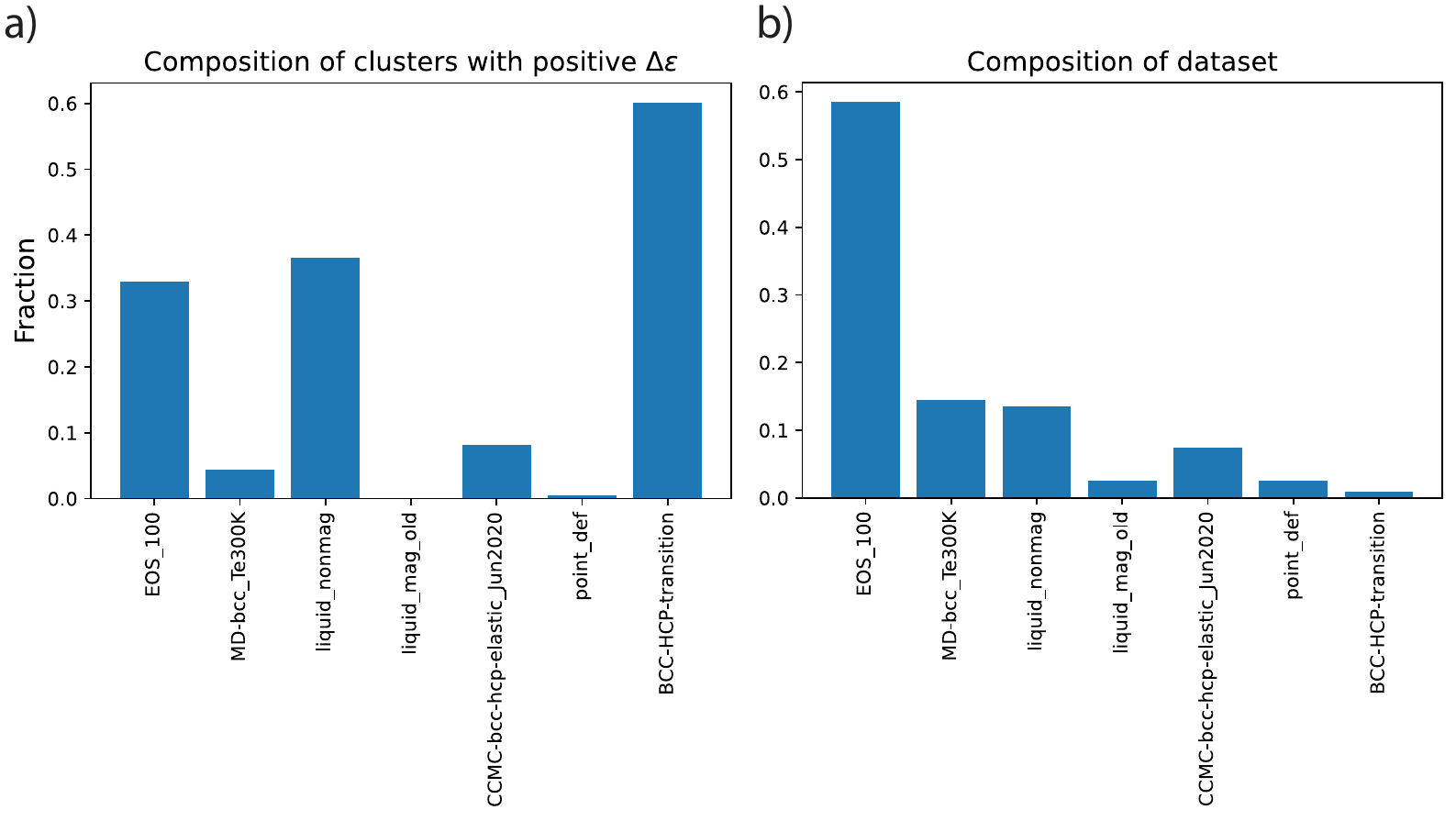}
    \end{center}
    \caption{(a) The fraction of atoms in each human-labeled group of the Fe dataset that are in clusters with positive $\Delta \epsilon$Fig.~\ref{fig:pruning_Fe}(b)] varies greatly by group and does not correlate with (b) the size of each group in the total dataset. By comparing the distributions in (a) and (b), we notice, e.g, that while the \grouplabel{BCC-HCP-transition} group makes up less than 1\% of the total dataset, nearly 60\% of the atoms in that group led to worse test errors when pruned, indicating that they are difficult training points to learn and could benefit from higher sampling.
     }
    \label{fig:compositions}
\end{figure}

The resulting pruning curves are shown in Figure~\ref{fig:pruning_Fe}, where we observe that while most clusters can be safely pruned (resulting in monotonically decreasing test errors upon pruning), some clusters lead to noticeably worse test errors when pruned. Since the test errors vary nearly linearly with pruning rate, we define a metric $\Delta \epsilon$ as the slope of the change in test error after pruning compared to the base MAE with respect to varying the pruning rate; a negative value of $\Delta \epsilon$ indicates improved model prediction.
We find that $\sim 66\%$ of the clusters have a negative $\Delta \epsilon$ value, while only two clusters have a value larger than 0.002.
Figure~\ref{fig:compositions} breaks down the composition of clusters with positive $\Delta \epsilon$ values (degraded model performance), revealing that large fractions of the \grouplabel{EOS\_100}, \grouplabel{liquid\_nonmag}, and \grouplabel{BCC-HCP-transition} groups lead to worse test errors when pruned, suggesting that they are difficult training points to learn and could benefit from increased sampling.
Upon inspection of the clusters with positive $\Delta \epsilon$ values, we find no clear correlation between the cluster size and the $\Delta \epsilon$ value, suggesting that further analysis is necessary to assess the degree to which different clusters can be pruned.
We hypothesize that incorporating an uncertainty quantification metric with MEAGraph would help to avoid pruning clusters that would increase the model error by only removing points from clusters with low uncertainty.

\section{Conclusion}

In conclusion, we have developed the MEAGraph model to address two critical challenges in dataset analysis for training force fields: (i) accurate identification of distinct atomic environments and (ii) efficient pruning of complex atomic datasets to obtain compact, yet diverse datasets. Our approach can significantly advance the development of machine learning potentials needed for various applications in chemistry, physics, and materials science. The results demonstrate the superior capability of the MEAGraph formalism in identifying intricate atomic environments, particularly in systems with high structural complexity, such as those containing dislocations or other defect structures. MEAGraph is particularly effective in this regard compared to other established clustering methods, including {\it k}-means and DBSCAN, showcasing the model's versatility in applications such as phase identification and defect analysis. Furthermore, we demonstrated the robustness of MEAGraph for pruning atomic datasets by applying the methodology on two publicly available datasets of elemental Ta  and Fe. 
Importantly, our approach enables unsupervised and unlabeled clustering, which eliminates the need for label information during training set optimization. 
Contrasting with some other clustering approaches, monotonic improvement in trained models and little stochastic variability of results is observed using MEAGraph.
From detailed analysis of the pruning performance of MEAGraph,
we observe that the majority of MEAGraph clusters can be pruned without hurting model performance, though additional work analyzing the effect of sample difficulty and model uncertainty are warranted.
Further generalization of the MEAGraph model framework---even beyond atomic datasets---allows it to be applied to any high-dimensional feature representation, making it adaptable for a wide range of applications, including clustering-based outlier detection, anomaly detection, and effective data pruning in high-dimensional feature spaces.

\section*{Author Contributions}

H.S. and V.L. conceptualized the study. V.L. supervised the research. H.S. developed the MEAGraph model, implemented the code, performed model training and analysis. J.V. performed model training and analysis. All authors contributed to discussions, data analysis, and manuscript preparation.

\section*{Acknowledgment}

This work was performed under the auspices of the U.S. Department of Energy by Lawrence Livermore National Laboratory under Contract DE-AC52-07NA27344. This work was funded by the Laboratory Directed Research and Development (LDRD) Program at LLNL under project tracking code 23-SI-006. Computing support for this work came from the Lawrence Livermore National Laboratory institutional computing facility. 

\newpage

\section{References}

\bibliographystyle{unsrt}
\bibliography{References}

\end{document}